# New Advances and Theoretical Insights into EDML

**Khaled S. Refaat** and **Arthur Choi** and **Adnan Darwiche**
Computer Science Department
University of California, Los Angeles
{krefaat,aychoi,darwiche}@cs.ucla.edu

## Abstract

EDML is a recently proposed algorithm for learning MAP parameters in Bayesian networks. In this paper, we present a number of new advances and insights on the EDML algorithm. First, we provide the multivalued extension of EDML, originally proposed for Bayesian networks over binary variables. Next, we identify a simplified characterization of EDML that further implies a simple fixed-point algorithm for the convex optimization problem that underlies it. This characterization further reveals a connection between EDML and EM: a fixed point of EDML is a fixed point of EM, and vice versa. We thus identify also a new characterization of EM fixed points, but in the semantics of EDML. Finally, we propose a hybrid EDML/EM algorithm that takes advantage of the improved empirical convergence behavior of EDML, while maintaining the monotonic improvement property of EM.

## 1 INTRODUCTION

EDML is a recently proposed algorithm for learning MAP parameters of a Bayesian network from incomplete data (Choi, Refaat, & Darwiche, 2011). EDML is procedurally very similar to Expectation Maximization (EM), yet EDML was shown to have certain advantages, both theoretically and practically. Theoretically, EDML can in certain specialized cases provably converge in one iteration, whereas EM may require many iterations to solve the same learning problem. Empirically, a preliminary experimental evaluation suggested that EDML could find better parameter estimates than EM, in fewer iterations.

In this paper, we present a number of new results and insights on the EDML algorithm. First, we provide a simple extension of EDML to Bayesian networks over multivalued variables, whereas EDML was initially proposed for Bayesian networks over binary variables. We also show that the convex optimization problem that underlies the binary version of EDML, remains convex for the multivalued case.

Next, we identify a new and simplified characterization of EDML, which facilitates a number of theoretical observations about EDML. For example, this new characterization implies a simple, fixed-point iterative algorithm for solving the convex optimization problems underlying EDML. Moreover, we show that this fixed-point algorithm monotonically improves the solutions of these convex optimization problems, which correspond to an approximate factorization of the posterior over network parameters.

Armed with this new characterization of EDML, we go on to identify a surprising connection between EDML and EM (considering their theoretical and practical differences). In particular, we show that a fixed point of EDML is a fixed point of EM, and vice versa. This observation has a number of implications. First, it provides a new perspective on EM fixed points, based on the semantics of EDML, which was originally inspired by an approximate inference algorithm for Bayesian networks that subsumed the influential loopy belief propagation algorithm as a degenerate case (Pearl, 1988; Choi & Darwiche, 2006). Second, it suggests a hybrid EDML/EM algorithm that seeks to take advantage of the desirable properties from each: the improved convergence behavior of EDML, and the monotonic improvement property of EM.

## 2 TECHNICAL PRELIMINARIES

We use upper case letters ($X$) to denote variables and lower case letters ($x$) to denote their values. Variable sets are denoted by bold-face upper case letters ($\mathbf{X}$) and their instantiations by bold-face lower case letters ($\mathbf{x}$). Generally, we will use $X$ to denote a variable in a Bayesian network and $\mathbf{U}$ to denote its parents. A network parameter will therefore have the general form $\theta_{x|\mathbf{u}}$, representing the probability $Pr(X=x|\mathbf{U}=\mathbf{u})$.

Each variable $X$ in a Bayesian network can be thought of as inducing a number of conditional random variables, denoted by $X|\mathbf{u}$, where the values of variable $X|\mathbf{u}$ are drawn based on the conditional distribution $Pr(X|\mathbf{u})$. Parameter estimation in Bayesian networks can be thought of as a process of estimating the distributions of these conditional random variables.

We will use $\theta$ to denote the set of all network parameters. Given a network structure $G$, our goal is to learn its parameters from an incomplete dataset, such as:

| example | $E$ | $B$ | $A$ | $C$ |
|---|---|---|---|---|
| 1 | $e_1$ | $b_1$ | $a_1$ | ? |
| 2 | ? | $b_2$ | $a_2$ | ? |
| 3 | $e_1$ | $b_2$ | $a_2$ | $c_1$ |

We use $\mathcal{D}$ to denote a dataset, and $\mathbf{d}_i$ to denote an example. The dataset above has three examples, with example $\mathbf{d}_2$ being $B = b_2$, and $A = a_2$.

### 2.1 LEARNING PARAMETERS

A commonly used measure for the quality of parameter estimates $\theta$ is their likelihood, defined as:

$$L(\theta|\mathcal{D}) = \prod_{i=1}^{N} Pr_\theta(\mathbf{d}_i),$$

where $Pr_\theta$ is the distribution induced by network structure $G$ and parameters $\theta$. In the case of complete data (each example fixes the value of each variable), the maximum likelihood (ML) parameters are unique and easily obtainable. Learning ML parameters is harder when the data is incomplete and the EM algorithm (Dempster, Laird, & Rubin, 1977; Lauritzen, 1995) is typically employed. EM starts with some initial parameters $\theta^0$, called a *seed,* and successively improves on them via iteration. EM uses the update equation:

$$\theta_{x|\mathbf{u}}^{k+1} = \frac{\sum_{i=1}^{N} Pr_{\theta^k}(x\mathbf{u}|\mathbf{d}_i)}{\sum_{i=1}^{N} Pr_{\theta^k}(\mathbf{u}|\mathbf{d}_i)},$$

which requires inference on a Bayesian network parameterized by $\theta^k$, in order to compute $Pr_{\theta^k}(x\mathbf{u}|\mathbf{d}_i)$ and $Pr_{\theta^k}(\mathbf{u}|\mathbf{d}_i)$. It is known that one run of the jointree algorithm on each example is sufficient to implement an iteration of EM, which is guaranteed to never decrease the likelihood of its estimates across iterations. EM also converges to every local maxima, given that it starts with an appropriate seed. It is common to run EM with multiple seeds, keeping the best local maxima it finds. See (Darwiche, 2009; Koller & Friedman, 2009) for recent treatments on parameter learning in Bayesian networks via EM and related methods.

EM can also be used to find Maximum a Posteriori (MAP) parameters given Dirichlet priors on network parameters. The Dirichlet prior for the parameters of a random variable $X|\mathbf{u}$ is specified by a set of exponents, $\psi_{x|\mathbf{u}}$, leading to a density $\propto \prod_x [\theta_{x|\mathbf{u}}]^{\psi_{x|\mathbf{u}}-1}$. It is common to assume that exponents are $> 1$, which guarantees a unimodal density. For MAP parameters, EM uses the update (see, e.g., Darwiche (2009)):

$$\theta_{x|\mathbf{u}}^{k+1} = \frac{\psi_{x|\mathbf{u}} - 1 + \sum_{i=1}^{N} Pr_{\theta^k}(x\mathbf{u}|\mathbf{d}_i)}{\psi_{X|\mathbf{u}} - |X| + \sum_{i=1}^{N} Pr_{\theta^k}(\mathbf{u}|\mathbf{d}_i)}, \qquad (1)$$

where $\psi_{X|\mathbf{u}} = \sum_x \psi_{x|\mathbf{u}}$. When $\psi_{x|\mathbf{u}} = 1$, the equation reduces to the one for computing ML parameters. Moreover, using $\psi_{x|\mathbf{u}} = 2$ leads to ML parameters with Laplace smoothing. This is a common technique to deal with the problem of insufficient counts (i.e., instantiations that never appear in the dataset, leading to zero probabilities and division by zero). We will use Laplace smoothing in our experiments.

### 2.2 SOFT EVIDENCE

EDML makes heavy use of soft evidence (i.e., evidence that changes the distribution of a variable without necessarily fixing its value). In this section, we give an introduction to the semantics of soft evidence.

We follow the treatment of (Chan & Darwiche, 2005) for soft evidence, which models soft evidence as hard evidence on a virtual event $\eta$. In particular, soft evidence on some variable $X$ with $k$ values is quantified by a vector $\lambda_{x_1}, \ldots, \lambda_{x_k}$ with $\lambda_{x_i} \in [0, \infty)$. The semantics is that $\lambda_{x_1} : \cdots : \lambda_{x_k} = Pr(\eta|x_1) : \cdots : Pr(\eta|x_k)$. The soft evidence on variable $X$ is then emulated by asserting the hard evidence $\eta$. That is, the new distribution on variable $X$ after having asserted the soft evidence is modeled by $Pr(X|\eta)$. Note that $Pr(X|\eta)$ depends only on the ratios $\lambda_{x_1} : \cdots : \lambda_{x_k}$, not on their absolute values. Hard evidence of the form $X = x_j$ can be modeled using $\lambda_{x_i} = 0$ for all $i \neq j$, and $\lambda_{x_j} = 1$. Moreover, neutral evidence can be modeled using $\lambda_{x_i} = 1$ for all $i$. The reader is referred to (Chan & Darwiche, 2005) for more details.

## 3 BINARY EDML

EDML is a recent method for learning Bayesian network parameters from incomplete data (Choi et al., 2011). It is based on Bayesian learning in which one formulates estimation in terms of computing posterior distributions on network parameters. That is, given a Bayesian network, one constructs a corresponding *meta network* in which parameters are explicated as variables, and on which the given dataset $\mathcal{D}$ can be asserted as evidence; see Figure 1. One then estimates parameters by considering the posterior distribution obtained from conditioning the meta network on the given dataset $\mathcal{D}$. Suppose for example that the meta network induces distribution $\mathcal{P}$ and let $\theta$ denote an instantiation of variables that represent parameters in the meta network. One can then obtain MAP parameter estimates by computing $\text{argmax}_\theta \, \mathcal{P}(\theta|\mathcal{D})$ using inference on the meta network; see (Darwiche, 2009) for an example treatment of Bayesian learning.

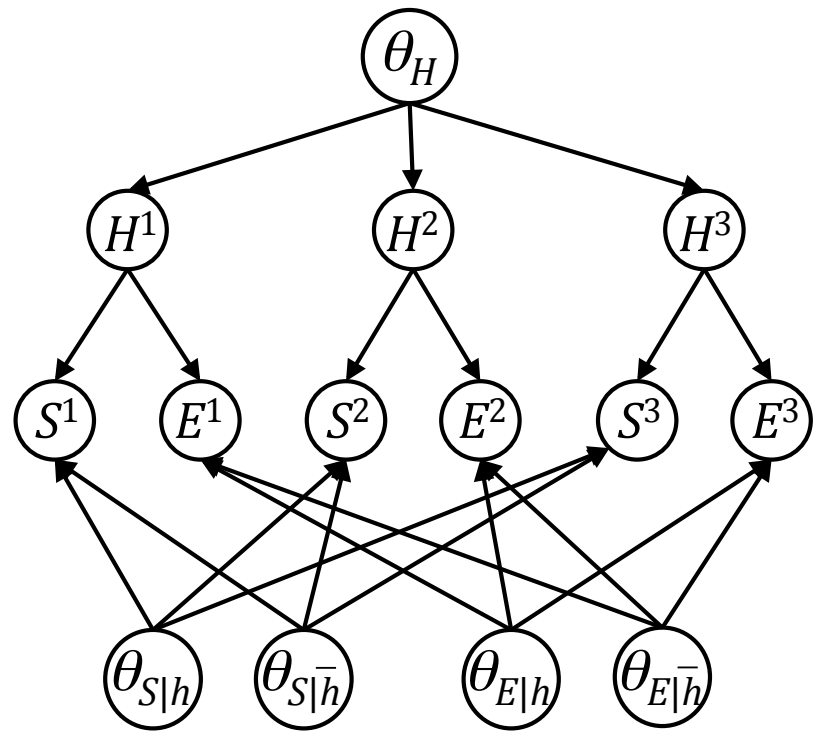

Figure 1: A meta network induced from a base network $S \longleftarrow H \longrightarrow E$. The CPTs here are based on standard semantics; see, e.g., (Darwiche, 2009, Ch. 18).

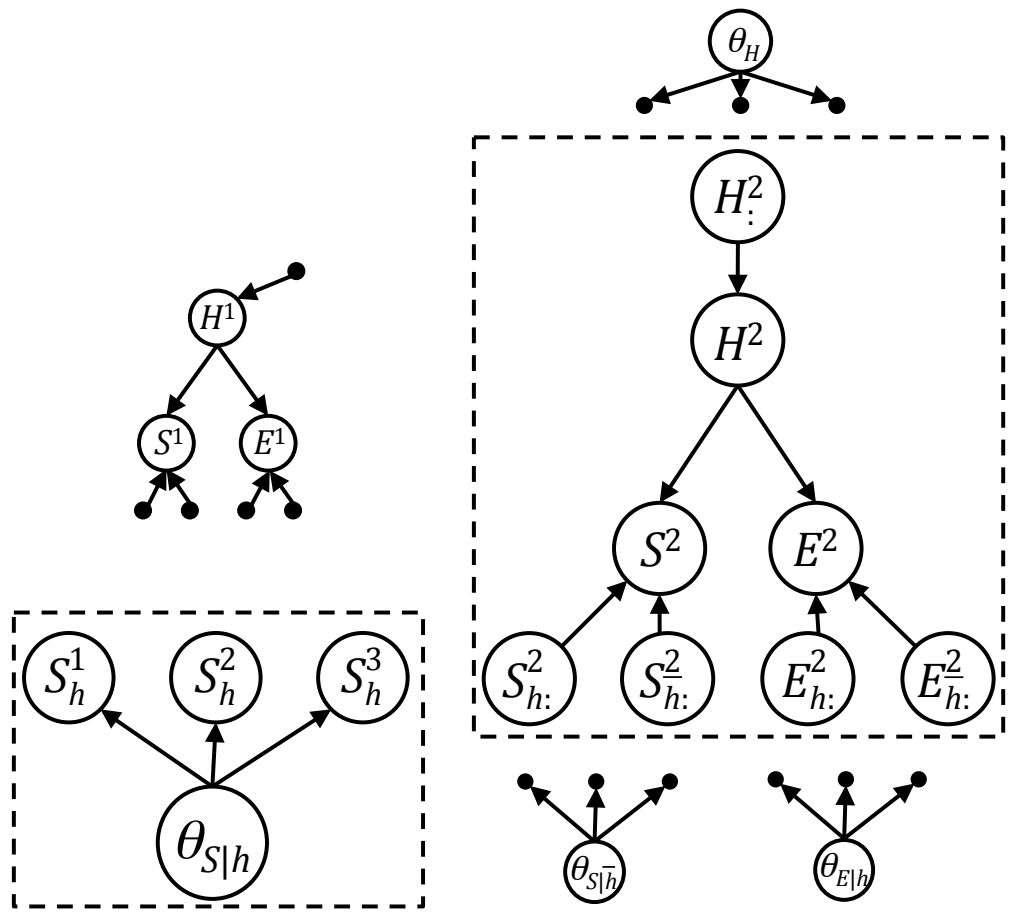

Figure 2: An edge-deleted network obtained from the meta network in Figure 1. Highlighted are the island for example $\mathbf{d}_2$ and the island for parameter set $\theta_{S|h}$.

It is known that meta networks tend to be too complex for exact inference algorithms, especially when the dataset is large enough. The basic insight behind EDML was to adapt a specific approximate inference scheme to meta networks with the goal of computing MAP parameter estimates. In particular, the original derivation of EDML adapted the approximate inference algorithm proposed by (Choi & Darwiche, 2006), in which edges are deleted from a Bayesian network

**Algorithm 1** Binary EDML

**input:**
- $G$: A Bayesian network structure
- $\mathcal{D}$: An incomplete dataset $\mathbf{d}_1, \ldots, \mathbf{d}_N$
- $\theta$: An initial parameterization of structure $G$
- $\alpha_{X|\mathbf{u}}, \beta_{X|\mathbf{u}}$: Beta prior for each random variable $X|\mathbf{u}$

1: **while** not converged **do**
2: $\quad Pr \leftarrow$ distribution induced by $\theta$ and $G$
3: $\quad$ **Compute** Bayes factors:

$$\kappa^i_{x|\mathbf{u}} \leftarrow \frac{Pr(x\mathbf{u}|\mathbf{d}_i)/Pr(x|\mathbf{u}) - Pr(\mathbf{u}|\mathbf{d}_i) + 1}{Pr(\bar{x}\mathbf{u}|\mathbf{d}_i)/Pr(\bar{x}|\mathbf{u}) - Pr(\mathbf{u}|\mathbf{d}_i) + 1}$$

$\quad$ for each family instantiation $x\mathbf{u}$ and example $\mathbf{d}_i$
4: $\quad$ **Update** parameters:

$$\theta_{x|\mathbf{u}} \leftarrow \underset{p}{\text{argmax}} \; [p]^{\alpha_{X|\mathbf{u}}-1}[1-p]^{\beta_{X|\mathbf{u}}-1} \prod_{i=1}^{N} [\kappa^i_{x|\mathbf{u}} \cdot p - p + 1]$$

5: **return** parameterization $\theta$

to make it sparse enough for exact inference, followed by a compensation scheme that attempts to improve the quality of the approximations obtained from the edge-deleted network. The adaptation of this inference method to meta networks is shown in Figure 2. The two specific techniques employed here were to augment each edge $\theta_{X|\mathbf{u}} \longrightarrow X^i$ by an auxiliary variable $X^i_{\mathbf{u}}$, leading to $\theta_{X|\mathbf{u}} \longrightarrow X^i_{\mathbf{u}} \longrightarrow X^i$, where $X^i_{\mathbf{u}} \longrightarrow X^i$ is an equivalence edge. This is followed by deleting the equivalence edge. This technique yielded a disconnected meta network with two classes of subnetworks, called *parameter islands* and *network islands.*

Deleting edges, as proposed by (Choi & Darwiche, 2006), leads to introducing two auxiliary nodes in the Bayesian network for each deleted edge. Moreover, approximate inference by edge deletion follows the deletion process by a compensation scheme that searches for appropriate CPTs of these auxiliary nodes. As it turns out, the search for these CPTs, which is done iteratively, was amenable to a very intuitive interpretation as shown in (Choi et al., 2011).

In particular, one set of CPTs corresponded to soft evidence on network parameters, where each network island contributes one piece of soft evidence for each network parameter. The second set of CPTs corresponded to updated parameter estimates, where each parameter island contributes an estimate of its underlying parameter set. This interpretation was the basis for the form of EDML shown in Algorithm 1. This particular version of EDML, introduced in (Choi et al., 2011), assumes that all network variables are binary. Binary EDML, as we shall call it, iterates just like EM does, producing new estimates after each iteration. However, EDML iterations can be viewed as having two phases. In the first phase, each example in the data set is used to compute a piece of soft evidence on each parameter set (Line 3 of Algorithm 1). In the second phase, the pieces of soft evidence pertaining to each parameter set are used to compute a new estimate of that set (by solving the convex optimization problem on Line 4 of Algorithm 1). The process repeats until some convergence criteria is met. Aside from this optimization task, EM and EDML have the same computational complexity.

# 4 MULTIVALUED EDML

One contribution of this paper is the extension of binary EDML so that it handles multivalued variables as well. In principle, the extension turns out to be straightforward and is depicted in Algorithm 3. However, two issues require further discussion. The first concerns the specification of soft evidence for multivalued variables. The second is confirming that the optimization problem corresponding to a parameter island (on Line 4 of Algorithm 3) remains strictly concave, therefore, admitting unique solutions. We will consider both issues next.

## 4.1 EXAMPLES AS SOFT EVIDENCE

The first key concept of EDML is to interpret a data example $\mathbf{d}_i$ in the dataset as soft evidence on a conditional random variable $X|\mathbf{u}$. As mentioned earlier, soft evidence on a variable is modeled using a vector of parameters, one for each value of the variable. We will therefore use $\lambda_{x|\mathbf{u}}$ to denote the parameter pertaining to value $x$ of variable $X|\mathbf{u}$.

EDML uses Equation 2 in Algorithm 3 to compute soft evidence. In particular, example $\mathbf{d}_i$ is viewed as soft evidence on conditional random variable $X|\mathbf{u}$ that is quantified as follows:

$$\lambda^i_{x|\mathbf{u}} \leftarrow Pr(x\mathbf{u}|\mathbf{d}_i)/Pr(x|\mathbf{u}) - Pr(\mathbf{u}|\mathbf{d}_i) + 1$$

We will not derive this equation here as it resembles the one for binary EDML. We will, however, discuss some of its key properties in this and further sections.

Consider the case when the example $\mathbf{d}_i$ is inconsistent with the parent instantiation $\mathbf{u}$. In this case, the example should be irrelevant to variable $X|\mathbf{u}$. Equation 2 does the right thing here as it reduces to 1 for all values of $x$, which amounts to neutral evidence.

Another special case is when example $\mathbf{d}_i$ is complete; that is, it has no missing values. In this case, one can verify that if $\mathbf{d}_i$ is consistent with $\mathbf{u}$ (relevant), then $\lambda^i_{x|\mathbf{u}} = 0$ for all values $x$ except the value $x^\star$ consistent with $\mathbf{d}_i$. This is equivalent to hard evidence in favor of $x^\star$. On the other hand, if $\mathbf{d}_i$ is inconsistent with $\mathbf{u}$ (irrelevant), then $\lambda^i_{x|\mathbf{u}} = 1$ for all values $x$, providing neutral evidence. In a nutshell, a complete example provides either hard evidence, if it is relevant; or neutral evidence, if it is irrelevant.

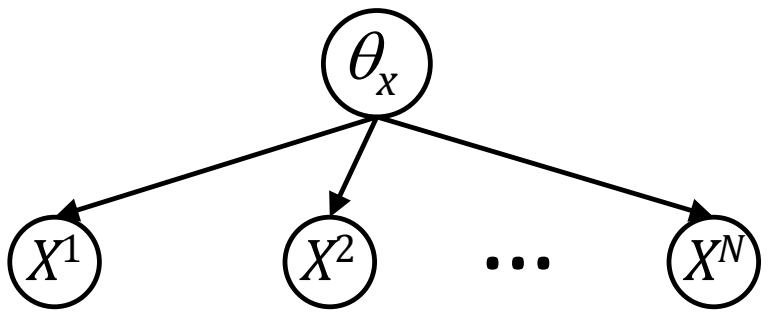

Figure 3: Learning from independent, hard observations $X^1, \ldots, X^N$. The distribution of variable $X$ is specified by parameter set $\theta_X$.

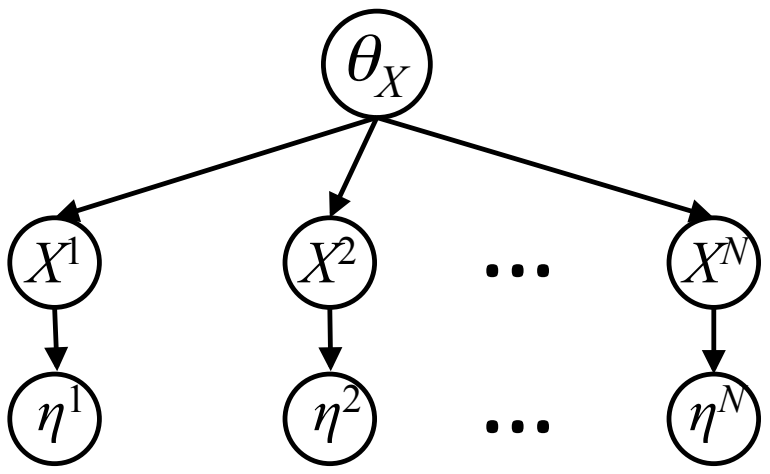

Figure 4: Learning from independent, soft observations $\eta_1, \ldots, \eta_N$. The distribution of variable $X$ is specified by parameter set $\theta_X$.

## 4.2 LEARNING FROM SOFT EVIDENCE

Consider the standard learning problem depicted in Figure 3. Here, we have a variable $X$ that takes $k$ values $x_1, \ldots, x_k$ and has a distribution specified by a parameter set $\theta_X : \theta_{x_1}, \ldots, \theta_{x_k}$. That is, each parameter $\theta_{x_i}$ represents the corresponding probability $Pr(X = x_i)$. Suppose further that we have a Dirichlet prior $\rho(\theta_X)$ on the parameter set with exponents $\psi_{x_i}$ greater than one. A standard learning problem here is to compute the MAP estimates of parameter set $\theta_X$ given $N$ independent observations on the variable $X$. MAP estimates are known to be unique in this case and have a corresponding closed form. In particular, it is known that the posterior $\rho(\theta_X|X^1, \ldots, X^N)$ is a unimodal Dirichlet and, hence, has a unique maximum; see for example (Darwiche, 2009).

EDML is based on a variant of this learning problem in which we are given $N$ *soft observations* on variable $X$ instead of hard observations. This variant is shown in Figure 4, where each observation $X^i$ has a child $\eta^i$ that is used to emulate soft evidence on $X^i$. That is, to represent soft evidence $\lambda^i_{x_1}, \ldots, \lambda^i_{x_k}$, we simply choose the CPT for $\eta^i$ so that $Pr(\eta^i|x_1) : \cdots : Pr(\eta^i|x_k) =$

**Algorithm 2** EM

**input:**
$G$: A Bayesian network structure
$\mathcal{D}$: An incomplete dataset $\mathbf{d}_1, \ldots, \mathbf{d}_N$
$\theta$: An initial parameterization of structure $G$
$\psi$: A Dirichlet prior for each parameter set $\theta_{X|\mathbf{u}}$

1: **while** not converged **do**
2: $Pr \leftarrow$ distribution induced by $\theta$ and $G$
3: **Compute** probabilities:

$$Pr(x\mathbf{u}|\mathbf{d}_i) \quad \text{and} \quad Pr(\mathbf{u}|\mathbf{d}_i)$$

for each family instantiation $x\mathbf{u}$ and example $\mathbf{d}_i$
4: **Update** parameters:

$$\theta_{x|\mathbf{u}} \leftarrow \frac{\psi_{x|\mathbf{u}} - 1 + \sum_{i=1}^N Pr(x\mathbf{u}|\mathbf{d}_i)}{\psi_{X|\mathbf{u}} - |X| + \sum_{i=1}^N Pr(\mathbf{u}|\mathbf{d}_i)}$$

5: **return** parameterization $\theta$

**Algorithm 3** Multivalued EDML

**input:**
$G$: A Bayesian network structure
$\mathcal{D}$: An incomplete dataset $\mathbf{d}_1, \ldots, \mathbf{d}_N$
$\theta$: An initial parameterization of structure $G$
$\psi$: A Dirichlet prior for each parameter set $\theta_{X|\mathbf{u}}$

1: **while** not converged **do**
2: $Pr \leftarrow$ distribution induced by $\theta$ and $G$
3: **Compute** soft evidence parameters:

$$\lambda^i_{x|\mathbf{u}} \leftarrow Pr(x\mathbf{u}|\mathbf{d}_i)/Pr(x|\mathbf{u}) - Pr(\mathbf{u}|\mathbf{d}_i) + 1 \quad (2)$$

for each family instantiation $x\mathbf{u}$ and example $\mathbf{d}_i$
4: **Update** parameters:

$$\theta_{X|\mathbf{u}} \leftarrow \operatorname*{argmax}_{\hat{\theta}_{X|\mathbf{u}}} \prod_x [\hat{\theta}_{x|\mathbf{u}}]^{\psi_{x|\mathbf{u}}-1} \prod_{i=1}^N \sum_x \lambda^i_{x|\mathbf{u}} \hat{\theta}_{x|\mathbf{u}} \quad (3)$$

5: **return** parameterization $\theta$

$\lambda^i_{x_1} : \cdots : \lambda^i_{x_k}$.

Note here that the posterior density $\rho(\theta_X|\eta^1 \ldots \eta^N)$ is no longer Dirichlet, but takes the more complex form given by Equation 3 of Algorithm 3. Yet, we have the following result.

**Theorem 1** *Given $N$ soft observations $\eta^i$ on a variable $X$, and a Dirichlet prior on its parameters $\theta_X$, with Dirichlet exponents $\psi_x > 1$, the posterior density $\rho(\theta_X|\eta^1 \ldots \eta^N)$ is strictly log concave.*

Therefore, the posterior density, $\rho(\theta_X|\eta^1 \ldots \eta^N)$, has a unique maximum. We next provide a simple iterative method for obtaining the maximum in this case.

## 5 SIMPLE EDML

Multivalued EDML, as given in Algorithm 3, works as follows. We start with some initial parameter estimates, just like EM. We then iterate, while performing two steps in each iteration. In the first step, each example $\mathbf{d}_i$ in the dataset is used to compute soft evidence on each variable $X|\mathbf{u}$, as in Equation 2 of Algorithm 3. Using this soft evidence, a learning problem is set up for the parameter set $\theta_{X|\mathbf{u}}$ as given in Figure 4, which is a learning *sub*-problem in the context of Equation 3 of Algorithm 3. The solution to this learning sub-problem provides the next estimate for parameter set $\theta_{X|\mathbf{u}}$. The process repeats. In principle, any appropriate optimization algorithm could be used to solve each of these learning sub-problems.

We will next derive a simpler description of EDML that has two key components. First, we will provide a convergent update equation that iteratively solves the optimization problem in Equation 3 of Algorithm 3. Second, we will use this update equation to provide a characterization of EDML's fixed points.

First, consider the likelihood:

$$Pr(\eta^1, \ldots, \eta^N \mid \theta_X) = \prod_{i=1}^N Pr(\eta^i \mid \theta_X)$$
$$= \prod_{i=1}^N \sum_x Pr(\eta^i \mid x) Pr(x \mid \theta_X) = \prod_{i=1}^N \sum_x \lambda^i_x \theta_x$$

Next, we have the posterior:

$$\rho(\theta_X \mid \eta^1, \ldots, \eta^N) \propto \rho(\theta_X) Pr(\eta^1, \ldots, \eta^N \mid \theta_X)$$
$$= \rho(\theta_X) \prod_{i=1}^N \sum_x \lambda^i_x \theta_x$$

Assuming a Dirichlet prior over parameter set $\theta_X$, with Dirichlet exponents $\psi_x$, the log of the posterior is:

$$\log \rho(\theta_X \mid \eta^1 \ldots \eta^N)$$
$$= \sum_x (\psi_x - 1) \log \theta_x + \sum_{i=1}^N \log \sum_x \lambda^i_x \theta_x + \gamma$$

where $\gamma$ is a constant that is independent of $\theta_X$, which we can ignore. By Theorem 1 we know that the log of the posterior is strictly concave when $\psi_x > 1$.

To get the unique maximum of the log posterior, we solve the optimization problem:

$$\begin{aligned} &\text{minimize} && -\log \rho(\theta_X \mid \eta^1, \ldots, \eta^N) \\ &\text{subject to} && \sum_x \theta_x = 1 \end{aligned}$$

In order to characterize the unique maximum of the log posterior, we start by taking the Lagrangian:

$$L(\theta_X, \nu) = -\log \rho(\theta_X \mid \eta^1, \ldots, \eta^N) + \nu \cdot (\sum_x \theta_x - 1)$$

where $\nu$ is a Lagrange multiplier. We get the following condition for the optimal parameter estimates, by setting the gradient of $L(\theta_X, \nu)$ with respect to $\theta_X$ to zero:

$$\theta_x = \frac{\psi_x - 1 + \sum_{i=1}^N \frac{\lambda_x^i \theta_x}{\sum_{x^*} \lambda_{x^*}^i \theta_{x^*}}}{\psi_X - |X| + N}$$

where $\psi_X = \sum_x \psi_x$, and where we used the constraint $\sum_x \theta_x = 1$ to identify that $\nu = \psi_X - |X| + N$.

The above equation leads to a much stronger result.

**Theorem 2** *The following update equation monotonically increases the posterior $\rho(\theta_{X|\mathbf{u}} \mid \eta^1, \ldots, \eta^N)$:*

$$\theta_{x|\mathbf{u}}^t = \frac{\psi_{x|\mathbf{u}} - 1 + \sum_{i=1}^N \frac{\lambda_{x|\mathbf{u}}^i \theta_{x|\mathbf{u}}^{t-1}}{\sum_{x^\star} \lambda_{x^\star|\mathbf{u}}^i \theta_{x^\star|\mathbf{u}}^{t-1}}}{\psi_{X|\mathbf{u}} - |X| + N} \quad (4)$$

This theorem suggests a convergent iterative algorithm for solving the convex optimization problem of Equation 3 in Algorithm 3. First, we start with some initial parameter estimates $\theta_{x|\mathbf{u}}^0$ at iteration $t = 0$. For iteration $t > 0$, we use the above update to compute parameters $\theta_{x|\mathbf{u}}^t$ given the parameters $\theta_{x|\mathbf{u}}^{t-1}$ from the previous iteration. If at some point, the parameters of one iteration do not change in the next (in practice, up to some limit), we say that the iterations have converged to a fixed point. The above theorem, together with Theorem 1, shows that these updates are convergent to the unique maximum of the posterior.

Given Equation 4, one can think of two types of *iterations* in EDML: local and global. A *global* iteration corresponds to executing Lines 2–4 of Algorithm 3 and is similar to an EM iteration. Within each global iteration, we have *local* iterations which correspond to the evaluations of Equation 4. Note that each parameter set $\theta_{X|\mathbf{u}}$ has its own local iterations, which are meant to find the optimal values of this parameter set. Moreover, the number of local iterations for each parameter set $\theta_{X|\mathbf{u}}$ may be different, depending on the soft evidence pertaining to that set (i.e., $\lambda_{x|\mathbf{u}}^i$) and depending on how Equation 4 is seeded for that particular parameter set (i.e., $\theta_{x|\mathbf{u}}^0$).

This leads to a number of observations on the difference between EM updates (Equation 1) and EDML updates (Equation 4). From a time complexity viewpoint, an EM update implies exactly one local iteration for each parameter set since Equation 1 needs to be evaluated only once for each parameter set. As mentioned earlier, however, an EDML update requires a varying number of local iterations. We have indeed observed that some parameter sets may require several hundred local iterations, depending on the seed of Equation 4 and the convergence criteria used. Another important observation is that EDML has a secondary set of seeds, as compared to EM, which are needed to start off Equation 4 at the beginning of each global iteration of EDML. In our experiments, we seed Equation 4 using the parameter estimates obtained from the previous global iteration of EDML. We note, however, that the choice of these secondary seeds is a subject that can significantly benefit from further research.

## 6 EDML FIXED POINTS

One of the more well known facts about EM is that its fixed points are precisely the stationary points of the log-likelihood function (or more generally, the posterior density when Dirichlet priors are used). This property has a number of implications, one of which is that EM is capable of converging to every local maxima of the log-likelihood, assuming that the algorithm is seeded appropriately.

In this section, we show that the fixed points of EDML are precisely the fixed points of EM. We start by formally defining what a fixed point is.

Both EM and EDML can be viewed as functions $f(\theta)$ that take a network parameterization $\theta$ and returns another network parameterization $f(\theta)$. Each algorithm is seeded with initial parameters $\theta^0$. After the first iteration, each algorithm produces the next parameters $\theta^1 = f(\theta^0)$. More generally, at iteration $i$, each algorithm produces the parameters $\theta^{i+1} = f(\theta^i)$. When $\theta^{i+1} = \theta^i$, we say that parameters $\theta^i$ are a fixed point for an algorithm. We also say that the algorithm has converged to $\theta^i$. We now have the following results.

**Theorem 3** *A parameterization $\theta$ is a fixed point for EDML if and only if it is a fixed point for EM.*

The proof of this theorem rests on two observations. First, using the update equation of EM given on Line 4 of Algorithm 2, one immediately gets that the EM fixed points are characterized by the following equation

$$Pr(x|\mathbf{u}) = \frac{\psi_{x|\mathbf{u}} - 1 + \sum_{i=1}^N Pr(x\mathbf{u}|\mathbf{d}_i)}{\psi_{X|\mathbf{u}} - |X| + \sum_{i=1}^N Pr(\mathbf{u}|\mathbf{d}_i)} \quad (5)$$

That is, we can test whether a network parameterization $\theta$ is a fixed point for EM by simply checking the probability distribution $Pr$ it induces to see if satisfies the above equation (this is actually a set of equations, one for each family instantiation $x\mathbf{u}$ in the network).

Consider now EDML updates as given by Equation 4 and suppose that we have reached a fixed point, where $\theta^t_{x|\mathbf{u}} = \theta^{t-1}_{x|\mathbf{u}} = Pr(x|\mathbf{u})$ for all $x\mathbf{u}$. If we now replace each $\lambda^i_{x|\mathbf{u}}$ by its corresponding value in Equation 2 of Algorithm 3, we obtain, after some simplification:

$$Pr(x|\mathbf{u}) = \frac{\psi_{x|\mathbf{u}} - 1 + \sum_{i=1}^{N} \frac{\lambda^i_{x|\mathbf{u}} Pr(x|\mathbf{u})}{\sum_{x^\star} \lambda^i_{x^\star|\mathbf{u}} Pr(x^\star|\mathbf{u})}}{\psi_{X|\mathbf{u}} - |X| + N}$$
$$= \frac{\psi_{x|\mathbf{u}} - 1 + \sum_{i=1}^{N} Pr(x\mathbf{u}|\mathbf{d}_i) + Pr(\neg\mathbf{u}|\mathbf{d}_i) Pr(x|\mathbf{u})}{\psi_{X|\mathbf{u}} - |X| + N} \qquad (6)$$

Thus, a parameterization $\theta$ is a fixed point for EDML iff it satisfies Equation 6. After noting that $N = \sum_{i=1}^{N} Pr(\mathbf{u}|\mathbf{d}_i) + Pr(\neg\mathbf{u}|\mathbf{d}_i)$, we rearrange Equation 6 to obtain Equation 5, which is the characteristic equation for EM fixed points. Thus, a parameterization $\theta$ satisfies Equation 6 iff it is a fixed point for EM.

# 7 HYBRID EDML/EM

By Theorem 3, EDML and EM share the same fixed points. This result is fairly surprising, considering the theoretical and practical differences between the two algorithms. For example, certain specialized situations were identified where EDML converges to optimal parameter estimates in a single global iteration (regardless of how it is seeded), whereas EM may require many iterations to converge to the same estimates (Choi et al., 2011). On the other hand, EDML is not guaranteed to monotonically improve its estimates after each global iteration as EM does (improvement here is in terms of increasing the likelihood of estimates, or the MAP when Dirichlet priors are used).

By carefully examining EDML updates (Equation 4), one can see that the quantities it needs to perform a single local iteration are the same quantities needed to perform an EM update (Equation 1). Hence, without any additional computational effort, one can obtain EM updates as a side effect of computing EDML updates (the converse is not true since Equation 4 may require many local iterations). As both algorithms share the same fixed points, it thus makes sense to consider a hybrid algorithm that takes advantage of the improved theoretical and practical benefits of EDML (with respect to faster convergence) and the monotonic improvement property of EM.

We thus propose a very simple hybrid algorithm. At each global iteration of EDML, we also compute EM updates simultaneously. We then evaluate each update and choose the one that increases the posterior the most. As EM is guaranteed to improve the posterior, this hybrid algorithm is also trivially guaranteed to monotonically improve the posterior, therefore, inheriting the most celebrated feature of EM.

While additionally computing an EM update is not much overhead if one is computing an EDML update, evaluating both updates with respect to the posterior incurs a non-trivial cost. As we shall see in the following section, however, the improved convergence behavior of EDML enables this hybrid algorithm to realize improvements over EM, both in terms of faster convergence (i.e., number of global iterations) and in terms of time (i.e., when taking both global and local iterations under consideration).

# 8 EXPERIMENTAL RESULTS

In our first set of experiments, we show that simple EDML, as compared to EM, can often find better estimates in fewer global iterations. In our second set of experiments, we show that a hybrid EDML/EM algorithm can find better estimates in less time, as well as fewer global iterations. We use the following networks: alarm, andes, asia, diagnose, pigs, spect, water, and win95pts. Network spect is a naive Bayes network induced from a dataset in the UCI ML repository, with 1 class variable and 22 attributes. Network diagnose is from the UAI 2008 evaluation. The other networks are commonly used benchmarks.[1]

Using these networks, we simulated data sets of a certain size ($2^{10}$), then made the data incomplete by randomly selecting a certain percentage of the nodes to be hidden (10%, 25%, 35%, 50%, and 70%). For each of these cases, and for each network, we further generated 3 data sets at random. A combination of a network, a percentage of hidden nodes, and a generated data set constitutes a learning problem. Both EM and EDML are seeded with the same set of randomly generated parameters. Local EDML iterations are seeded with the estimates of the previous global iteration.

## 8.1 EXPERIMENTS I

First, we study the behavior of EDML compared to EM, with respect to global iterations (more on the computation time, in the next section). For every learning problem, we run both EM and EDML[2] for 1000 global iterations and identify the best MAP estimates achieved by either of them, for the purpose of evaluation. In each global iteration, EM and EDML each try to improve their current estimates by improving the posterior (again, EM is provably guaranteed to increase the posterior, while EDML is not). The difference between the (log) posterior of the current estimates, and the best (log) posterior found, by either algorithm, is considered to be the error. The error is measured at every global iteration of EM and EDML until it decreases below $10^{-4}$. Table 1 summarizes the results by showing the percentage of global iterations in which each algorithm had less error than the other. In the global iterations where an algorithm had less error, the factor by which it decreases the error of the other algorithm, on average, is computed, and is considered the relative improvement: $r$ and $r'$, for EM and EDML, respectively. Table 1 shows the results for three different breakdowns: (1) by different networks, (2) by different hiding percentages, and (3) on average.

[1] Available at `http://www.cs.huji.ac.il/site/labs/compbio/Repository/` and `http://genie.sis.pitt.edu/networks.html`

[2] Soft evidence and parameter updates are damped in EDML, which is typical for algorithms like loopy belief propagation, which EDML is, in part, inspired by (Choi & Darwiche, 2006).

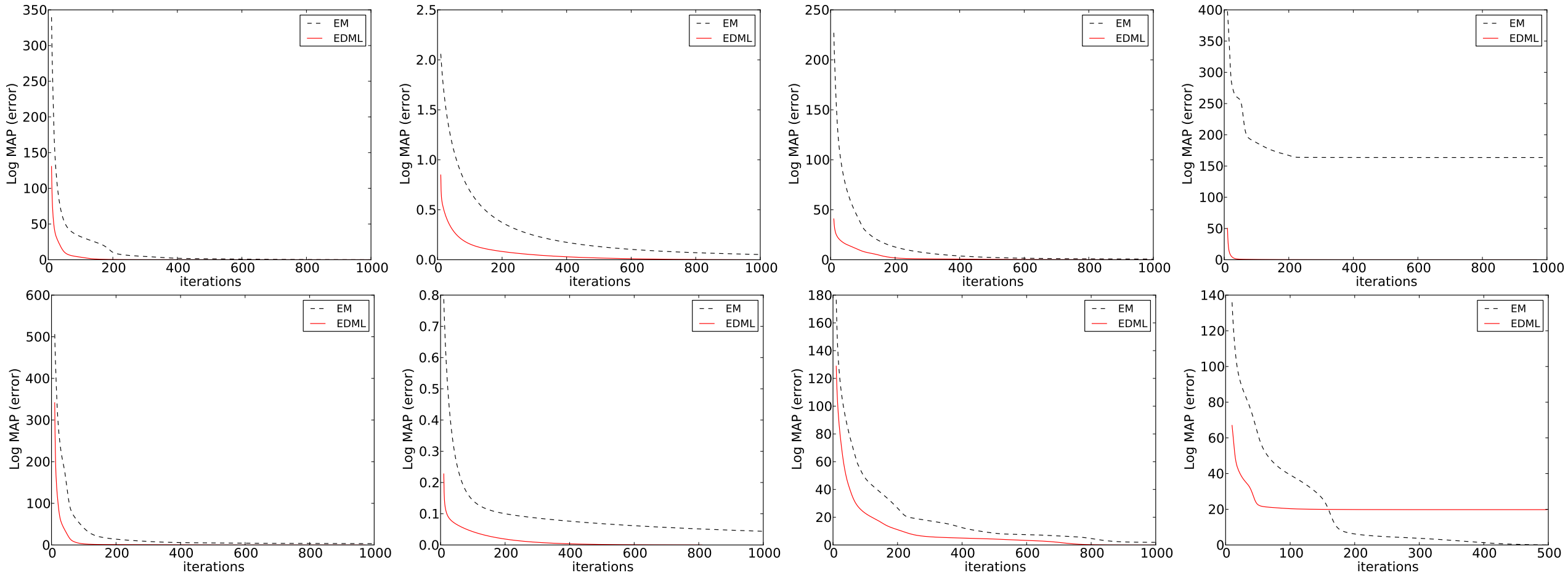

Figure 5: MAP Error of parameter estimates over iterations. Going right on the $x$-axis, we have increasing iterations. Going up on the $y$-axis, we have increasing error. EDML is depicted with a solid red line, and EM with a dashed black line. The curves are from left to right, in pairs, for the networks: andes, asia, diagnose, and alarm. Each pair of curves represents a selection of two different datasets of size $2^{10}$.

Table 1: Speedup results (iterations)

| category | % EDML | % EM | $r$ | $r'$ |
|---|---|---|---|---|
| alarm | 89.25% | 10.75% | 76.21% | 76.44% |
| andes | 75.89% | 24.11% | 88.95% | 79.29% |
| asia | 99.01% | 0.99% | 92.05% | 76.91% |
| diagnose | 78.99% | 21.01% | 77.99% | 80.18% |
| pigs | 83.34% | 16.66% | 83.51% | 60.57% |
| spect | 86.65% | 13.35% | 82.70% | 79.96% |
| water | 82.77% | 17.23% | 91.55% | 83.78% |
| win95pts | 78.73% | 21.27% | 91.75% | 79.89% |
| **hiding 10%** | 93.82% | 6.18% | 84.59% | 87.13% |
| **hiding 25%** | 90.95% | 9.05% | 83.83% | 75.70% |
| **hiding 35%** | 82.24% | 17.76% | 86.26% | 75.09% |
| **hiding 50%** | 77.61% | 22.39% | 87.8% | 80.21% |
| **hiding 70%** | 75.65% | 24.35% | 84.48% | 74.21% |
| **average** | 83.05% | 16.95% | 85.41% | 76.96% |

We see here that EDML can obtain better estimates than EM in much fewer global iterations. Interestingly, this is the case even though EDML is not guaranteed to improve estimates after each global iteration, as EM does. Another interesting observation is that decreasing the percentage of hidden nodes widens the gap between EDML and EM, in favor of EDML. This is not surprising though since the approximate inference scheme on which EDML is based becomes more accurate with more observations. In particular, the local optimization problems that EDML solves exactly and independently, become more independent with more observations (i.e., as the dataset becomes more complete). Figure 5 highlights a selection of error curves given by EM and EDML for different learning problems. One can see that in most cases shown, the EDML error goes to zero much faster than EM.

## 8.2 EXPERIMENTS II

Our first set of experiments showed that EDML can obtain better estimates in significantly fewer global iterations than EM. A global EDML iteration, however, is more costly than a global EM iteration as EDML performs local iterations which are needed to solve the convex optimization problem associated with each parameter set. Thus, EDML can potentially take more time to converge than EM in some cases, therefore reducing the overall benefit over EM, in terms of time (more on this later).[3] EDML can still perform favorably time-wise compared to EM, but we show here

[3]This could be alleviated, for example, by better seeding of the local EDML iterations (to speed up convergence of the local iterations), or by performing the local EDML iterations in parallel, across parameter sets.

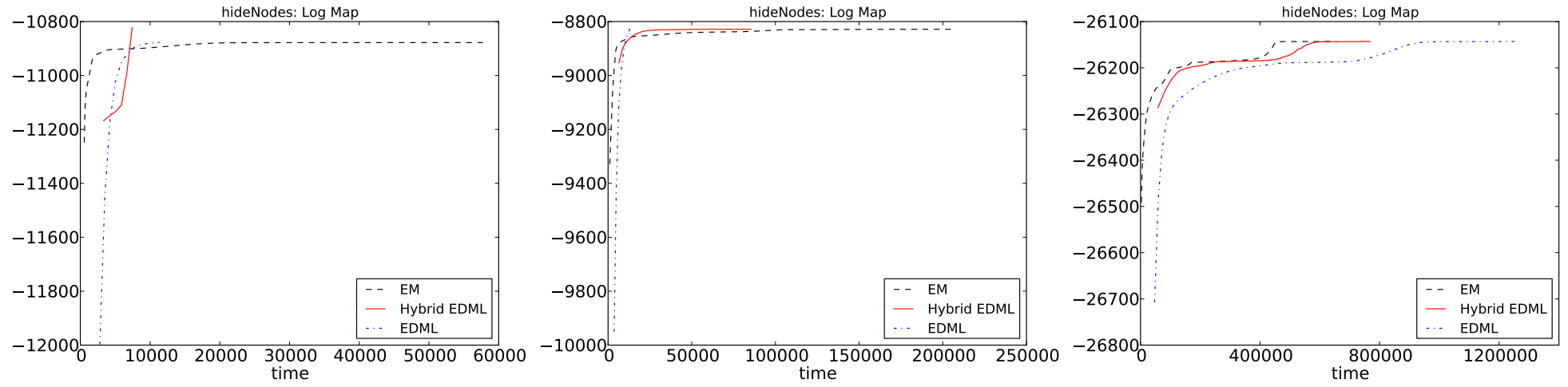

Figure 6: MAP of parameter estimates over time. Going right on the $x$-axis, we have increasing time (ms). Going up on the $y$-axis, we have increasing MAP. Hybrid EDML is depicted with a solid red line, EM with a dashed black line, and EDML with a blue dotted dashed line. The curves are from left to right for the following problems: alarm with hiding 25%, win95pts with hiding 35%, and water with hiding 50%.

Table 2: Speedup results (time)

| network | % Hybrid | % EM | $s$ | $s'$ |
|---|---|---|---|---|
| alarm | 46.67% | 53.33% | 75.80% | 56.22% |
| andes | 53.33% | 46.67% | 42.27% | 47.08% |
| asia | 66.67% | 33.33% | 70.37% | 41.00% |
| diagnose | 26.67% | 73.33% | 52.71% | 43.14% |
| pigs | 73.33% | 26.67% | 52.35% | 31.32% |
| spect | 100% | 0% | 98.03% | — |
| water | 35.71% | 64.29% | 46.15% | 43.55% |
| win95pts | 80.00% | 20.00% | 66.37% | 54.95% |
| **average** | 60.50% | 39.50% | 67.45% | 45.55% |

that a hybrid EDML/EM can go even further.

We first run EM until convergence (or until it exceeds 1000 iterations). Second, we run our hybrid EDML/EM until it achieves the same quality of parameter estimates as EM (or until it exceeds 1000 iterations). To summarize the results, Table 2 shows the percentage of learning problems in which each algorithm was faster than the other. In the cases where the hybrid EDML/EM algorithm was faster, the average percentage by which it decreases the execution time of EM is reported as the speedup ($s$). The speedup for the cases in which EM was faster is given by $s'$.

The results suggest that hybrid EDML/EM can be used to get better estimates in less time. Specifically, on average, the hybrid method decreases the execution time by a factor of 3.07 in about 60.50% of the cases, and, in the rest of the cases, EM was faster by a factor of roughly 1.84. This is particularly interesting in light of the non-trivial overhead associated with hybrid EDML/EM, as it has to evaluate both EDML and EM estimates in order to select the better estimate. In comparison, EDML alone was on average faster than EM by a factor of 2.59 times in about 54.17% of the cases, whereas EM was faster than EDML alone by a factor of 2.38 in the remaining 45.83% of the cases.

Figure 6 shows a sample of the time curves showing two cases where EDML/EM reached better MAP estimates, faster than EM, and one case where EDML/EM finished slightly after EM. EDML alone is also plotted in Figure 6 where it is usually slower than EDML/EM except in some cases; see the center plot in Figure 6.

# 9 RELATED WORK

EM has played a critical role in learning probabilistic graphical models and Bayesian networks (Dempster et al., 1977; Lauritzen, 1995; Heckerman, 1998). However, learning (and Bayesian learning in particular) remains challenging in a variety of situations, particularly when there are hidden (latent) variables; see, e.g., (Elidan, Ninio, Friedman, & Shuurmans, 2002; Elidan & Friedman, 2005). More recently, there have been characterizations of EM that also have interesting connections to loopy belief propagation, and related algorithms (Liu & Ihler, 2011; Jiang, Rai, & III, 2011), although their focus is more on approximate (partial) MAP inference in probabilistic graphical models via message-passing, and less on learning.

Slow convergence of EM has also been recognized, particularly in the presence of hidden variables. In such cases, EM can be coupled with algorithms such as gradient ascent, or other more traditional algorithms for optimization; see, e.g. (Aitkin & Aitkin, 1996). A variety of other techniques for accelerating EM have been proposed in the literature; see, e.g., (Thiesson, Meek, & Heckerman, 2001).

### Acknowledgements

This work has been partially supported by ONR grant #N00014-12-1-0423, NSF grant #IIS-1118122, and NSF grant #IIS-0916161. This work is also based upon research performed in collaborative facilities renovated with funds from NSF grant #0963183, an award funded under the American Recovery and Reinvestment Act of 2009 (ARRA).